\pdfoutput=1

\documentclass[11pt]{article}

\usepackage{acl}

\usepackage{times}
\usepackage{latexsym}

\usepackage[T1]{fontenc}

\usepackage[utf8]{inputenc}

\usepackage{microtype}

\usepackage{caption}
\usepackage{subcaption}
\usepackage[ruled,vlined]{algorithm2e}
\usepackage{smile}

\newcommand{\ours}{\text{MoEBERT~}}

\usepackage{setspace}
\AtBeginDocument{%
  \addtolength\abovedisplayskip{-0.35\baselineskip}%
  \addtolength\belowdisplayskip{-0.35\baselineskip}%
  \addtolength\abovedisplayshortskip{-0.35\baselineskip}%
  \addtolength\belowdisplayshortskip{-0.35\baselineskip}%
}

\title{MoEBERT: from BERT to Mixture-of-Experts via \\Importance-Guided Adaptation}

\author{
Simiao Zuo$^\ddagger$\thanks{\hspace{0.03in} Corresponding author.}, \ Qingru Zhang$^\ddagger$, \
Chen Liang$^\ddagger$, \
Pengcheng He$^\diamond$, \\
\textbf{Tuo Zhao$^{\ddagger}$ and Weizhu Chen$^\diamond$} \\
$^\ddagger$Georgia Institute of Technology \ \
$^\diamond$Microsoft \\
\texttt{\{simiaozuo,qzhang441,cliang73,tourzhao\}@gatech.edu}\\
\texttt{\{Pengcheng.H,wzchen\}@microsoft.com}
}

\begin{document}
\maketitle

\begin{abstract}
Pre-trained language models have demonstrated superior performance in various natural language processing tasks. However, these models usually contain hundreds of millions of parameters, which limits their practicality because of latency requirements in real-world applications.
Existing methods train small compressed models via knowledge distillation.
However, performance of these small models drops significantly compared with the pre-trained models due to their reduced model capacity.
We propose MoEBERT, which uses a Mixture-of-Experts structure to increase model capacity and inference speed.
We initialize MoEBERT by adapting the feed-forward neural networks in a pre-trained model into multiple experts. As such, representation power of the pre-trained model is largely retained.
During inference, only one of the experts is activated, such that speed can be improved.
We also propose a layer-wise distillation method to train MoEBERT.
We validate the efficiency and effectiveness of MoEBERT on natural language understanding and question answering tasks.
Results show that the proposed method outperforms existing task-specific distillation algorithms. For example, our method outperforms previous approaches by over $2\%$ on the MNLI (mismatched) dataset.
Our code is publicly available at \url{https://github.com/SimiaoZuo/MoEBERT}.
\end{abstract}

\section{Introduction}

Pre-trained language models have demonstrated superior performance in various natural language processing tasks, such as natural language understanding \citep{devlin2018bert, liu2019roberta, he2020deberta} and natural language generation \citep{radford2019language, brown2020language}.
These models can contain billions of parameters, e.g., T5 \citep{raffel2019exploring} contains up to $11$ billion parameters, and GPT-3 \citep{brown2020language} consists of up to $175$ billion parameters.
Their extreme sizes bring challenges in serving the models to real-world applications due to latency requirements.

Model compression through knowledge distillation \citep{romero2014fitnets, hinton2015distilling} is a promising approach that reduces the computational overhead of pre-trained language models while maintaining their superior performance.
In knowledge distillation, a large pre-trained language model serves as a teacher, and a smaller student model is trained to mimic the teacher's behavior.
Distillation approaches can be categorized into two groups: task-agnostic \citep{sanh2019distilbert, jiao2019tinybert, wang2020minilm, wang2020minilmv2, sun2020contrastive} and task-specific \citep{turc2019well, sun2019patient, li2020bert, hou2020dynabert, sun2020mobilebert, xu2020bert}.
Task-agnostic distillation pre-trains the student and then fine-tunes it on downstream tasks; while task-specific distillation directly fine-tunes the student after initializing it from a pre-trained model.
Note that task-agnostic approaches are often combined with task-specific distillation during fine-tuning for better performance \citep{jiao2019tinybert}.
We focus on task-specific distillation in this work.

One major drawback of existing knowledge distillation approaches is the drop in model performance caused by the reduced representation power. That is, because the student model has fewer parameters than the teacher, its model capacity is smaller.
For example, the student model in DistilBERT \citep{sanh2019distilbert} has $66$ million parameters, about half the size of the teacher (BERT-base, \citealt{devlin2018bert}). Consequently, performance of DistilBERT drops significantly compared with BERT-base, e.g., over $2\%$ on MNLI ($82.2$ v.s. $84.5$) and over $3\%$ on CoLA ($54.7$ v.s. $51.3$).

We resort to the Mixture-of-Experts (MoE, \citealt{shazeer2017outrageously}) structure to remedy the representation power issue. MoE models can increase model capacity while keeping the inference computational cost constant.
A layer of a MoE model \citep{shazeer2017outrageously, lepikhin2020gshard, fedus2021switch, yang2021exploring, zuo2021taming} consists of an attention mechanism and multiple feed-forward neural networks (FFNs) in parallel. Each of the FFNs is called an expert. During training and inference, an input adaptively activates a fixed number of experts (usually one or two).
In this way, the computational cost of a MoE model remains constant during inference, regardless of the total number of experts.
Such a property facilitates compression without reducing model capacity.

However, MoE models are difficult to train-from-scratch and usually require a significant amount of parameters, e.g., $7.4$ billion parameters for Switch-base \citep{fedus2021switch}.
We propose MoEBERT, which incorporates the MoE structure into pre-trained language models for fine-tuning.
Our model can speedup inference while retaining the representation power of the pre-trained language model.
Specifically, we incorporate the expert structure by adapting the FFNs in a pre-trained model into multiple experts. For example, the hidden dimension of the FFN is $3072$ in BERT-base \citep{devlin2018bert}, and we adapt it into $4$ experts, each has a hidden dimension $768$. In this way, the amount of \emph{effective parameters} (i.e., parameters involved in computing the representation of an input) is cut by half, and we obtain a $\times 2$ speedup.
We remark that \ours utilizes more parameters of the pre-trained model than existing approaches, such that it has greater representation power.

To adapt the FFNs into experts, we propose an importance-based method. 
Empirically, there are some neurons in the FFNs that contribute more to the model performance than the other ones. That is, removing the important neurons causes significant performance drop. Such a property can be quantified by the \emph{importance score} \citep{molchanov2019importance, xiao2019autoprune, liang2021super}.
When initializing MoEBERT, we share the most important neurons (i.e., the ones with the highest scores) among the experts, and the other neurons are distributed evenly.
This strategy has two advantages:
first, the shared neurons preserve performance of the pre-trained model;
second, the non-shared neurons promote diversity among experts, which further boost model performance.
After initialization, \ours is trained using a layer-wise task-specific distillation algorithm.

We demonstrate efficiency and effectiveness of \ours on natural language understanding and question answering tasks. On the GLUE \citep{wang2018glue} benchmark, our method significantly outperforms existing distillation algorithms. For example, \ours exceeds performance of state-of-the-art task-specific distillation approaches by over $2\%$ on the MNLI (mismatched) dataset. For question answering, \ours increases F1 by $2.6$ on SQuAD v1.1 \citep{squad1} and $7.0$ on SQuAD v2.0 \citep{squad2} compared with existing algorithms.

The rest of the paper is organized as follows: we introduce background and related works in Section~\ref{sec:background}; we describe \ours in Section~\ref{sec:method}; experimental results are provided in Section~\ref{sec:experiment}; and Section~\ref{sec:conclusion} concludes the paper.
\section{Background}
\label{sec:background}

\subsection{Backbone: Transformer}

The Transformer \citep{vaswani2017attention} backbone has been widely adopted in pre-trained language models. The model contains several identically-constructed Transformer layers. Each layer has a multi-head self-attention mechanism and a two-layer feed-forward neural network (FFN).

Suppose the output of the attention mechanism is $\mathbf{A}$. Then, the FFN is defined as:
\begin{align} \label{eq:transformer-ffn}
    \mathbf{H} = \sigma ( \mathbf{A} \mathbf{W}_1 + \mathbf{b}_1 ), \
    \mathbf{X}^\ell = \mathbf{W}_2 \mathbf{H} + \mathbf{b}_2,
\end{align}
where $\mathbf{W}_1 \in \RR^{d \times d_h}$, $\mathbf{W}_2 \in \RR^{d_h \times d}$, $\mathbf{b}_1 \in \RR^{d_h}$ and $\mathbf{b}_2 \in \RR^d$ are weights of the FFN, and $\sigma$ is the activation function. Here $d$ denotes the embedding dimension, and $d_h$ denotes the hidden dimension of the FFN.

\subsection{Mixture-of-Experts Models}
\label{sec:intro-moe}

Mixture-of-Experts models consist of multiple expert layers, which are similar to the Transformer layers. Each of these layers contain a self-attention mechanism and multiple FFNs (Eq.~\ref{eq:transformer-ffn}) in parallel, where each FFN is called an expert.

Let $\{E_i\}_{i=1}^N$ denote the experts, and $N$ denotes the total number of experts. Similar to Eq.~\ref{eq:transformer-ffn}, the experts in layer $\ell$ take the attention output $\mathbf{A}$ as the input. For each $\mathbf{a}_t$ (the $t$-th row of $\mathbf{A}$) that corresponds to an input token, the corresponding output $\mathbf{x}_t^\ell$ of layer $\ell$ is 
\begin{align} \label{eq:moe-output}
    & \mathbf{x}_t^\ell = \sum_{i \in \cT} p_i(\mathbf{a}_t) E_i(\mathbf{a}_t).
\end{align}
Here, $\cT \subset \{1 \cdots N\}$ is the activated set of experts with $|\cT|=K$, and $p_i$ is the weight of expert $E_i$.

Different approaches have been proposed to construct $\cT$ and compute $p_i$. For example, \citet{shazeer2017outrageously} take
\begin{align} \label{eq:moe-gate}
    & p_i(\mathbf{a}_t) = \left[ \mathrm{softmax}\left(\mathbf{a}_t \mathbf{W}_g \right) \right]_i,
\end{align}
where $\mathbf{W}_g$ is a weight matrix. Consequently, $\cT$ is constructed as the experts that yield top-$K$ largest $p_i$.
However, such an approach suffers from load imbalance, i.e., $\mathbf{W}_g$ collapses such that nearly all the inputs are routed to the same expert. Existing works adopt various ad-hoc heuristics to mitigate this issue, e.g., adding Gaussian noise to Eq.~\ref{eq:moe-gate} \citep{shazeer2017outrageously}, limiting the maximum number of inputs that can be routed to an expert \citep{lepikhin2020gshard}, imposing a load balancing loss \citep{lepikhin2020gshard, fedus2021switch}, and using linear assignment \citep{lewis2021base}.
In contrast, \citealt{roller2021hash} completely remove the gate and pre-assign tokens to experts using hash functions, in which case we can take $p_i=1/K$.

In Eq.~\ref{eq:moe-output}, a token only activates $K$ instead of $N$ experts, and usually $K \ll N$, e.g., $K=2$ and $N=2048$ in GShard \citep{lepikhin2020gshard}.
As such, the number of FLOPs for one forward pass does not scale with the number of experts. Such a property paves the way for increasing inference speed of a pre-trained model without decreasing the model capacity, i.e., we can adapt the FFNs in a pre-trained model into several smaller components, and only activate one of the components for a specific input token.

\subsection{Pre-trained Language Models}
Pre-trained language models \citep{peters2018deep, devlin2018bert, raffel2019exploring, liu2019roberta, brown2020language, he2020deberta, he2021debertav3} have demonstrated superior performance in various natural language processing tasks.
These models are trained on an enormous amount of unlabeled data, such that they contain rich semantic information that benefits downstream tasks. 
Fine-tuning pre-trained language models achieves state-of-the-art performance in tasks such that natural language understanding \citep{he2021debertav3} and natural language generation \citep{brown2020language}.

\subsection{Knowledge Distillation}
Knowledge distillation \citep{romero2014fitnets, hinton2015distilling} compensates for the performance drop caused by model compression. In knowledge distillation, a small student model mimics the behavior of a large teacher model.
For example, DistilBERT \citep{sanh2019distilbert} uses the teacher's soft prediction probability to train the student model; TinyBERT \citep{jiao2019tinybert} aligns the student's layer outputs (including attention outputs and hidden states) with the teacher's; MiniLM \citep{wang2020minilm, wang2020minilmv2} utilizes self-attention distillation; and CoDIR \citep{sun2020contrastive} proposes to use a contrastive objective such that the student can distinguish positive samples from negative ones according to the teacher's outputs.

There are also heated discussions on the number of layers to distill. For example, \citet{wang2020minilm, wang2020minilmv2} distill the attention outputs of the last layer; \citet{sun2019patient} choose specific layers to distill; and \citet{jiao2019tinybert} use different weights for different transformer layers. 

There are two variants of knowledge distillation: task-agnostic \citep{sanh2019distilbert, jiao2019tinybert, wang2020minilm, wang2020minilmv2, sun2020contrastive} and task-specific \citep{turc2019well, sun2019patient, li2020bert, hou2020dynabert, sun2020mobilebert, xu2020bert}.
The former requires pre-training a small model using knowledge distillation and then fine-tuning on downstream tasks, while the latter directly fine-tunes the small model.
Note that task-agnostic approaches are often combined with task-specific distillation for better performance, e.g., TinyBERT \citep{jiao2019tinybert}.
In this work, we focus on task-specific distillation.

\section{Method}
\label{sec:method}

In this section, we first present an algorithm that adapts a pre-trained language model into a MoE model. Such a structure enables inference speedup by reducing the number of parameters involved in computing an input token's representation.
Then, we introduce a layer-wise task-specific distillation method that compensates for the performance drop caused by model compression.

\begin{figure}[t!]
    \centering
    \includegraphics[width=0.8\linewidth]{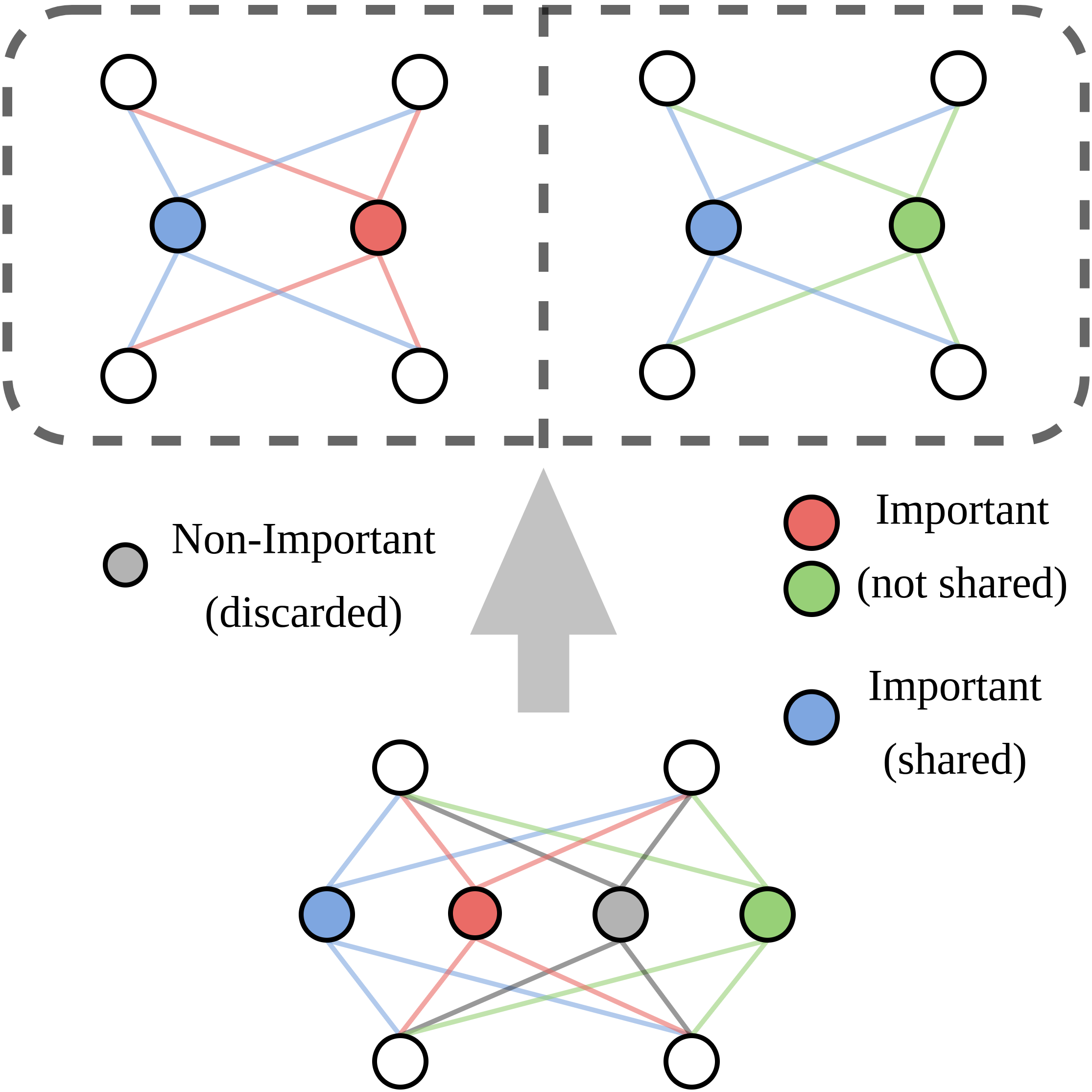}
    \caption{Adapting a two-layer FFN into two experts. The blue neuron is the most important one, and is shared between the two experts. The red and green neurons are the second and third important ones, and are assigned to expert one and two, respectively.}
    \label{fig:moebert-arch}
    \vskip -0.1in
\end{figure}

\subsection{Importance-Guided Adaptation of Pre-trained Language Models}
\label{sec:moe}

Adapting the FFNs in a pre-trained language model into multiple experts facilitates inference speedup while retaining model capacity. This is because in a MoE model, only a subset of parameters are used to compute the representation of a given token (Eq.~\ref{eq:moe-output}). These activated parameters are referred to as \emph{effective parameters}.
For example, by adapting the FFNs in a pre-trained BERT-base \citep{devlin2018bert} (with hidden dimension $3072$) model into $4$ experts (each has hidden dimension $768$), the number of effective parameters reduces by half, such that we obtain a $\times 2$ speedup.

Empirically, we find that randomly converting a FFN into experts works poorly (see Figure~\ref{fig:ablation:split-method} in the experiments). This is because there are some columns in $\mathbf{W}_1 \in \RR^{d \times d_h}$ (correspondingly some rows in $\mathbf{W}_2$ in Eq.~\ref{eq:transformer-ffn}) contribute more than the others to model performance.

The importance score \citep{molchanov2019importance, xiao2019autoprune, liang2021super}, originally introduced in model pruning literature, measures such parameter importance. For a dataset $\cD$ with sample pairs $\{(x,y)\}$, the score is defined as
\begin{align} \label{eq:importance}
    I_j = \sum_{(x,y) \in \cD} &\Big| (\mathbf{w}_j^1)^\top \nabla_{\mathbf{w}_j^1} \cL(x,y) \notag \\
    &+ (\mathbf{w}_j^2)^\top \nabla_{\mathbf{w}_j^2} \cL (x,y)
    \Big|.
\end{align}
Here $\mathbf{w}_j^1 \in \RR^d$ is the $j$-th column of $\mathbf{W}_1$, $\mathbf{w}_j^2$ is the $j$-th row of $\mathbf{W}_2$, and $\cL(x,y)$ is the loss.

The importance score in Eq.~\ref{eq:importance} indicates variation of the loss if we remove the neuron. That is,
\begin{align*}
    |\cL_{\mathbf{w}} - \cL_{\mathbf{w}=\mathbf{0}}| &\approx
    \left| (\mathbf{w}-\mathbf{0})^\top \nabla_{\mathbf{w}}\cL_{\mathbf{w}} \right| \\
    &= |\mathbf{w}^\top \nabla_{\mathbf{w}}\cL_{\mathbf{w}}|,
\end{align*}
where $\cL_{\mathbf{w}}$ is the loss with neuron\footnote{A neuron $\mathbf{w}$ contains two weights $\mathbf{w}^1$ and $\mathbf{w}^2$ as in Eq.~\ref{eq:importance}.} $\mathbf{w}$ and $\cL_{\mathbf{w}=\mathbf{0}}$ is the loss without neuron $\mathbf{w}$. Here the approximation is based on the first order Taylor expansion of $\cL_\mathbf{w}$ around $\mathbf{w}=\mathbf{0}$.

After computing $I_j$ for all the columns, we adapt $\mathbf{W}_1$ into experts.\footnote{The other parameters in the FFN: $\mathbf{W}_2$, $\mathbf{b}_1$ and $\mathbf{b}_2$ are treated similarly according to $\{I_j\}$.} The columns are sorted in ascending order according to their importance scores as $\mathbf{w}_{(1)}^1 \cdots \mathbf{w}_{(d_h)}^1$, where $\mathbf{w}_{(1)}^1$ has the largest $I_j$ and $\mathbf{w}_{(d_h)}^1$ the smallest.
Empirically, we find that sharing the most important columns benefits model performance. Based on this finding, suppose we share the top-$s$ columns and we adapt the FFN into $N$ experts, then expert $e$ contains columns $\{\mathbf{w}_{(1)}^1, \cdots, \mathbf{w}_{(s)}^1, \mathbf{w}_{(s+e)}^1, \mathbf{w}_{(s+e+N)}^1, \cdots \}$.
Note that we discard the least important columns to keep the size of each expert as $\lfloor d/N \rfloor$.
Figure~\ref{fig:moebert-arch} is an illustration of adapting a FFN with $4$ neurons in a pre-trained model into two experts.

\subsection{Layer-wise Distillation}

To remedy the performance drop caused by adapting a pre-trained model to a MoE model, we adopt a layer-wise task-specific distillation algorithm. We use BERT-base \citep{devlin2018bert} as both the student (i.e., the MoE model) and the teacher. We distill both the Transformer layer output $\mathbf{X}^\ell$ (Eq.~\ref{eq:moe-output}) and the final prediction probability.

For the Transformer layers, the distillation loss is the mean squared error between the teacher's layer output $\mathbf{X}^\ell_{\text{tea}}$ and the student's layer output $\mathbf{X}^\ell$ obtained from Eq.~\ref{eq:moe-output}.\footnote{Note that Eq.~\ref{eq:moe-output} computes the layer output of one token $\mathbf{x}_t^\ell$, i.e., one row in $\mathbf{X}^\ell$.}
Concretely, for an input $x$, the Transformer layer distillation loss is
\begin{align} \label{eq:trm-loss}
    \cL\textsubscript{trm}(x) =
    \sum_{\ell=0}^{L} \mathrm{MSE}(\mathbf{X}^\ell, \mathbf{X}_{\text{tea}}^\ell),
\end{align}
where $L$ is the total number of layers. Notice that we include the MSE loss of the embedding layer outputs $\mathbf{X}^0$ and $\mathbf{X}_{\text{tea}}^0$.

Let $f$ denotes the MoE model and $f_{\text{tea}}$ the teacher model. We obtain the prediction probability for an input $x$ as $p=f(x)$ and $p_{\text{tea}} = f_{\text{tea}}(x)$, where $p$ is the prediction of the MoE model and $p_{\text{tea}}$ is the prediction of the teacher model. Then the distillation loss for the prediction layer is
\begin{align} \label{eq:pred-loss}
    \cL\textsubscript{pred}(x) = \frac{1}{2} \left( \mathrm{KL}(p || p\textsubscript{tea})
    + \mathrm{KL}(p\textsubscript{tea} || p) \right),
\end{align}
where $\mathrm{KL}$ is the Kullback–Leibler divergence.

The layer-wise distillation loss is the sum of Eq.~\ref{eq:trm-loss} and Eq.~\ref{eq:pred-loss}, defined as
\begin{align} \label{eq:distil-loss}
    \cL\textsubscript{distill}(x) = \cL\textsubscript{trm}(x) + \cL\textsubscript{pred}(x).
\end{align}
We will discuss variants of Eq.~\ref{eq:distil-loss} in the experiments.


\subsection{Model Training}

We employ the random hashing strategy \citep{roller2021hash} to train the experts. That is, each token is pre-assigned to a random expert, and this assignment remains the same during training and inference. We will discuss more about other routing strategies of the MoE model in the experiments.

Given the training dataset $\cD$ and samples $\{(x,y)\}$, the training objective is
\begin{align*}
    \cL = \sum_{(x,y) \in \cD} \mathrm{CE}(f(x), y) + \lambda\textsubscript{distill} \cL\textsubscript{distill}(x),
\end{align*}
where $\mathrm{CE}$ is the cross-entropy loss and $\lambda\textsubscript{distill}$ is a hyper-parameter.


\section{Experiments}
\label{sec:experiment}

In this section, we evaluate the effectiveness and efficiency of the proposed algorithm on natural language understanding and question answering tasks. We implement our algorithm using the \textit{Huggingface Transformers}\footnote{\url{https://github.com/huggingface/transformers}} \citep{wolf2019huggingface} code-base. All the experiments are conducted on NVIDIA V100 GPUs.

\subsection{Datasets}

\textbf{GLUE.}
We evaluate performance of the proposed method on the General Language Understanding Evaluation (GLUE) benchmark \cite{wang2018glue}, which is a collection of nine natural language understanding tasks.
The benchmark includes two single-sentence classification tasks: SST-2 \citep{sst2013} is a binary classification task that classifies movie reviews to positive or negative, and CoLA \citep{cola2018} is a linguistic acceptability task. 
GLUE also contains three similarity and paraphrase tasks: MRPC \citep{mrpc2005} is a paraphrase detection task; STS-B \citep{sts-b2017} is a text similarity task; and QQP is a duplication detection task.
There are also four natural language inference tasks in GLUE: MNLI \citep{mnli2018}; QNLI \citep{squad1}; RTE \citep{rte1, rte2, rte3, rte5}; and WNLI \citep{wnli2012}.
Following previous works on model distillation, we exclude STS-B and WNLI in the experiments.
Dataset details are summarized in Appendix~\ref{app:dataset}.

\vspace{0.1in} \noindent
\textbf{Question Answering.}
We evaluate the proposed algorithm on two question answering datasets: SQuAD v1.1 \citep{squad1} and SQuAD v2.0 \citep{squad2}. These tasks are treated as a sequence labeling problem, where we predict the probability of each token being the start and end of the answer span.
Dataset details can be found in Appendix~\ref{app:dataset}.

\begin{table*}[t!]
\centering
\begin{tabular}{lccccccc}
\toprule
\multicolumn{1}{l|}{} & \textbf{RTE} & \textbf{CoLA} & \textbf{MRPC} & \textbf{SST-2} & \textbf{QNLI} & \textbf{QQP} & \textbf{MNLI} \\
\multicolumn{1}{l|}{} & Acc & Mcc & F1/Acc & Acc & Acc & F1/Acc & m/mm \\ \midrule
\multicolumn{1}{l|}{BERT-base} & 63.5 & 54.7 & 89.0/84.1 & 92.9 & 91.1 & 88.3/90.9 & 84.5/84.4 \\ \midrule
\textbf{Task-agnostic} &  &  &  &  &  &  &  \\
\multicolumn{1}{l|}{DistilBERT} & 59.9 & 51.3 & 87.5/- & 92.7 & 89.2 & -/88.5 & 82.2/- \\
\multicolumn{1}{l|}{TinyBERT (w/o aug)} & 72.2 & 42.8 & 88.4/- & 91.6 & 90.5 & -/90.6 & 83.5/- \\
\multicolumn{1}{l|}{MiniLMv1} & 71.5 & 49.2 & 88.4/- & 92.0 & 91.0 & -/91.0 & 84.0/- \\
\multicolumn{1}{l|}{MiniLMv2} & 72.1 & 52.5 & 88.9/- & 92.4 & 90.8 & -/91.1 & 84.2/- \\
\multicolumn{1}{l|}{CoDIR (pre+fine)} & 67.1 & 53.7 & 89.6/- & \textbf{93.6} & 90.1 & -/89.1 & 83.5/82.7 \\ \midrule
\textbf{Task-specific} &  &  &  &  &  &  &  \\
\multicolumn{1}{l|}{PKD} & 65.5 & 24.8 & 86.4/- & 92.0 & 89.0 & -/88.9 & 81.5/81.0 \\
\multicolumn{1}{l|}{BERT-of-Theseus} & 68.2 & 51.1 & 89.0/- & 91.5 & 89.5 & -/89.6 & 82.3/- \\
\multicolumn{1}{l|}{CoDIR (fine)} & 65.6 & 53.6 & 89.4/- & \textbf{93.6} & 90.4 & -/89.1 & 83.6/82.8 \\ \midrule
\textbf{Ours (task-specific)} &  &  &  &  &  &  &  \\
\multicolumn{1}{l|}{\ours} & \textbf{74.0} & \textbf{55.4} & \textbf{92.6/89.5} & 93.0 & \textbf{91.3} & \textbf{88.4/91.4} & \textbf{84.5/84.8} \\
\bottomrule
\end{tabular}
\caption{Experimental results on the GLUE development set. The best results are shown in \textbf{bold}. All the models are trained without data augmentation. All the models have $66M$ parameters, except BERT-base ($110M$ parameters). We report mean over three runs. Model references: BERT \citep{devlin2018bert}, DistilBERT \citep{sanh2019distilbert}, TinyBERT \citep{jiao2019tinybert}, MiniLMv1 \citep{wang2020minilm}, MiniLMv2 \citep{wang2020minilmv2}, CoDIR \citep{sun2020contrastive}, PKD \citep{sun2019patient}, BERT-of-Theseus \citep{xu2020bert}.}
\label{tab:glue-results}
\end{table*}

\subsection{Baselines}
We compare our method with both task-agnostic and task-specific distillation methods.

In task-agnostic distillation, we pre-train a small language model through knowledge distillation, and then fine-tune on downstream tasks. The fine-tuning procedure also incorporates task-specific distillation for better performance.

\vspace{0.05in} \noindent
\textbf{DistilBERT} \citep{sanh2019distilbert} pre-trains a small language model by distilling the temperature-controlled soft prediction probability.

\vspace{0.05in} \noindent
\textbf{TinyBERT} \citep{jiao2019tinybert} is a task-agnostic distillation method that adopts layer-wise distillation.

\vspace{0.05in} \noindent
\textbf{MiniLMv1} \citep{wang2020minilm} and \textbf{MiniLMv2} \citep{wang2020minilmv2} pre-train a small language model by aligning the attention distribution between the teacher model and the student model.

\vspace{0.05in} \noindent
\textbf{CoDIR} (Contrastive Distillation, \citealt{sun2020contrastive}) proposes a framework that distills knowledge through intermediate Transformer layers of the teacher via a contrastive objective.

\vspace{0.05in}
In task-specific distillation, a pre-trained language model is directly compressed and fine-tuned.

\vspace{0.05in} \noindent
\textbf{PKD} (Patient Knowledge Distillation, \citealt{sun2019patient}) proposes a method where the student patiently learns from multiple intermediate Transformer layers of the teacher.

\vspace{0.05in} \noindent
\textbf{BERT-of-Theseus} \citep{xu2020bert} proposes a progressive module replacing method for knowledge distillation.

\subsection{Implementation Details}
In the experiments, we use BERT-base \citep{devlin2018bert} as both the student model and the teacher model. That is, we first transform the pre-trained model into a MoE model, and then apply layer-wise task-specific knowledge distillation.
We set the number of experts in the MoE model to $4$, and the hidden dimension of each expert is set to $768$, a quarter of the hidden dimension of BERT-base. The other configurations remain unchanged. We share the top-$512$ important neurons among the experts (see Section~\ref{sec:moe}).
The number of effective parameters of the MoE model is $66M$ (v.s. $110M$ for BERT-base), which is the same as the baseline models.
We use the random hashing strategy \citep{roller2021hash} to train the MoE model, we will discuss more later.
Detailed training and hyper-parameter settings can be found in Appendix~\ref{app:training}.

\begin{table*}[t!]
\centering
\begin{tabular}{lcccc}
\toprule
\multicolumn{1}{l|}{} & \multicolumn{2}{c|}{\textbf{SQuAD v1.1}} & \multicolumn{2}{c}{\textbf{SQuAD v2.0}} \\
\multicolumn{1}{l|}{} & EM & \multicolumn{1}{c|}{F1} & EM & F1 \\ \midrule
\multicolumn{1}{l|}{BERT-base \citep{devlin2018bert}} & 80.7 & \multicolumn{1}{c|}{88.4} & 74.5 & 77.7 \\ \midrule
\textbf{Task-agnostic} &  &  &  &  \\
\multicolumn{1}{l|}{DistilBERT \citep{sanh2019distilbert}} & 78.1 & \multicolumn{1}{c|}{86.2} & 66.0 & 69.5 \\
\multicolumn{1}{l|}{TinyBERT (w/o aug) \citep{jiao2019tinybert}} & - & \multicolumn{1}{c|}{-} & - & 73.1 \\
\multicolumn{1}{l|}{MiniLMv1 \citep{wang2020minilm}} & - & \multicolumn{1}{c|}{-} & - & 76.4 \\
\multicolumn{1}{l|}{MiniLMv2 \citep{wang2020minilmv2}} & - & \multicolumn{1}{c|}{-} & - & 76.3 \\ \midrule
\textbf{Task-specific} &  &  &  &  \\
\multicolumn{1}{l|}{PKD \citep{sun2019patient}} & 77.1 & \multicolumn{1}{c|}{85.3} & 66.3 & 69.8 \\ \midrule
\textbf{Ours (task-specific)} &  &  &  &  \\
\multicolumn{1}{l|}{\ours} & \textbf{80.4} & \multicolumn{1}{c|}{\textbf{87.9}} & \textbf{73.6} & \textbf{76.8} \\
\bottomrule
\end{tabular}
\caption{Experimental results on SQuAD v1.1 and SQuAD v2.0. The best results are shown in \textbf{bold}. All the models are trained without data augmentation. All the models have $66M$ parameters, except BERT-base ($109M$ parameters). Here \textit{EM} means exact match.}
\label{tab:squad-results}
\end{table*}

\begin{figure*}[t!]
\centering
\begin{subfigure}{0.3\textwidth}
    \centering
    \includegraphics[width=1.0\textwidth]{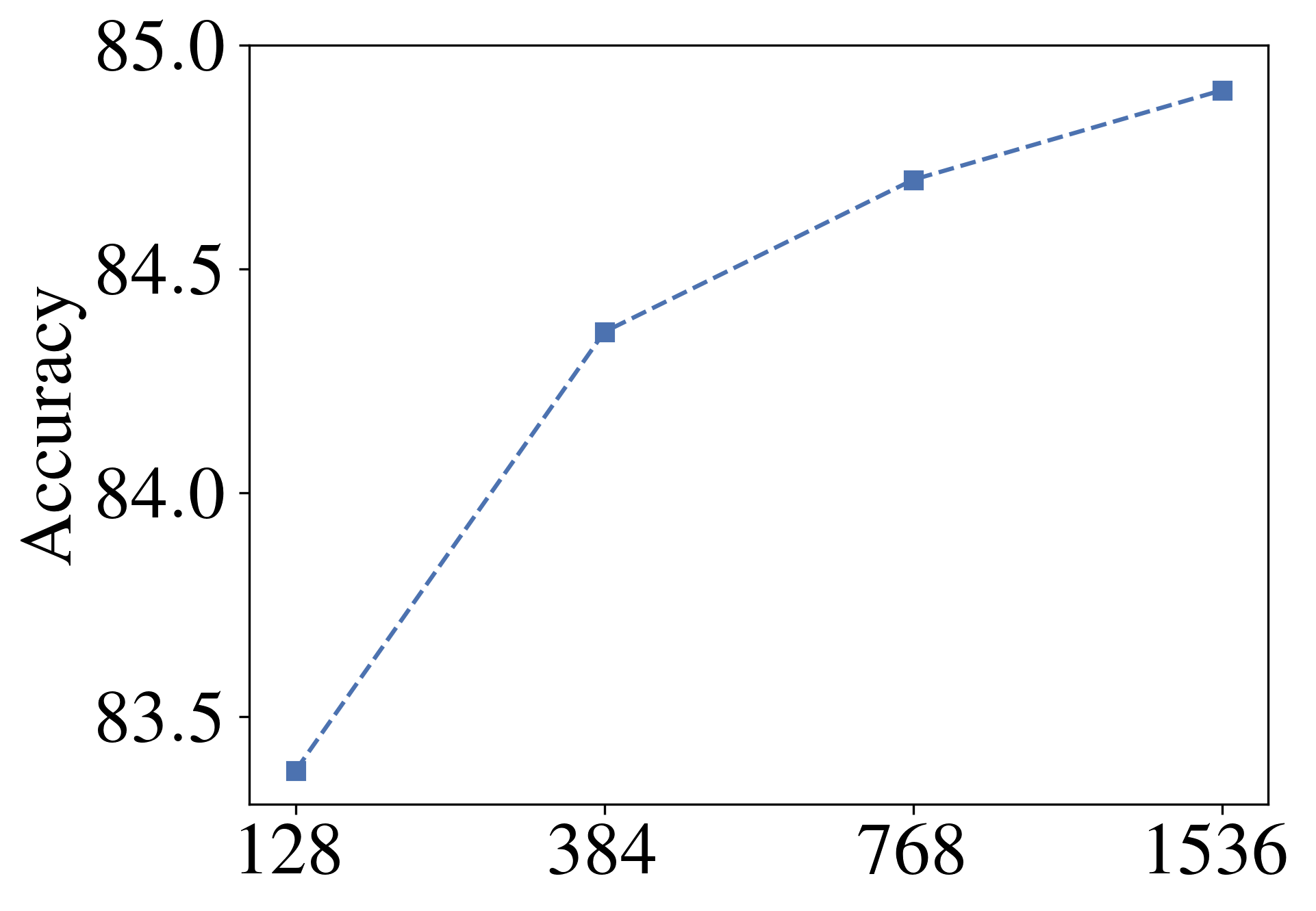}
    \caption{Expert dimension.}
    \label{fig:ablation:expert-dim}
\end{subfigure} \hspace{0.05in}
\begin{subfigure}{0.3\textwidth}
    \centering
    \includegraphics[width=1.0\textwidth]{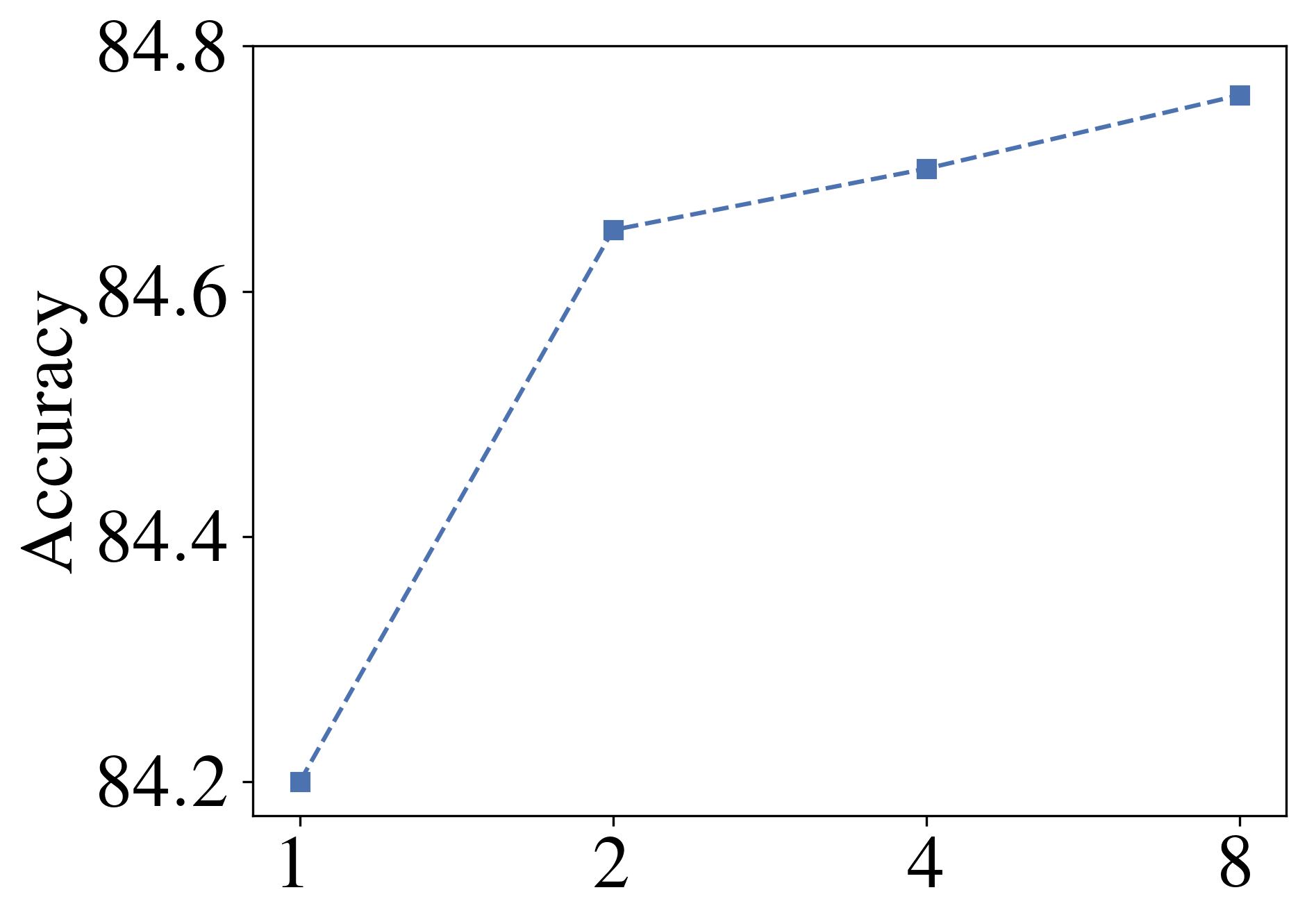}
    \caption{Number of experts.}
    \label{fig:ablation:num-experts}
\end{subfigure} \hspace{0.05in}
\begin{subfigure}{0.3\textwidth}
    \centering
    \includegraphics[width=1.0\textwidth]{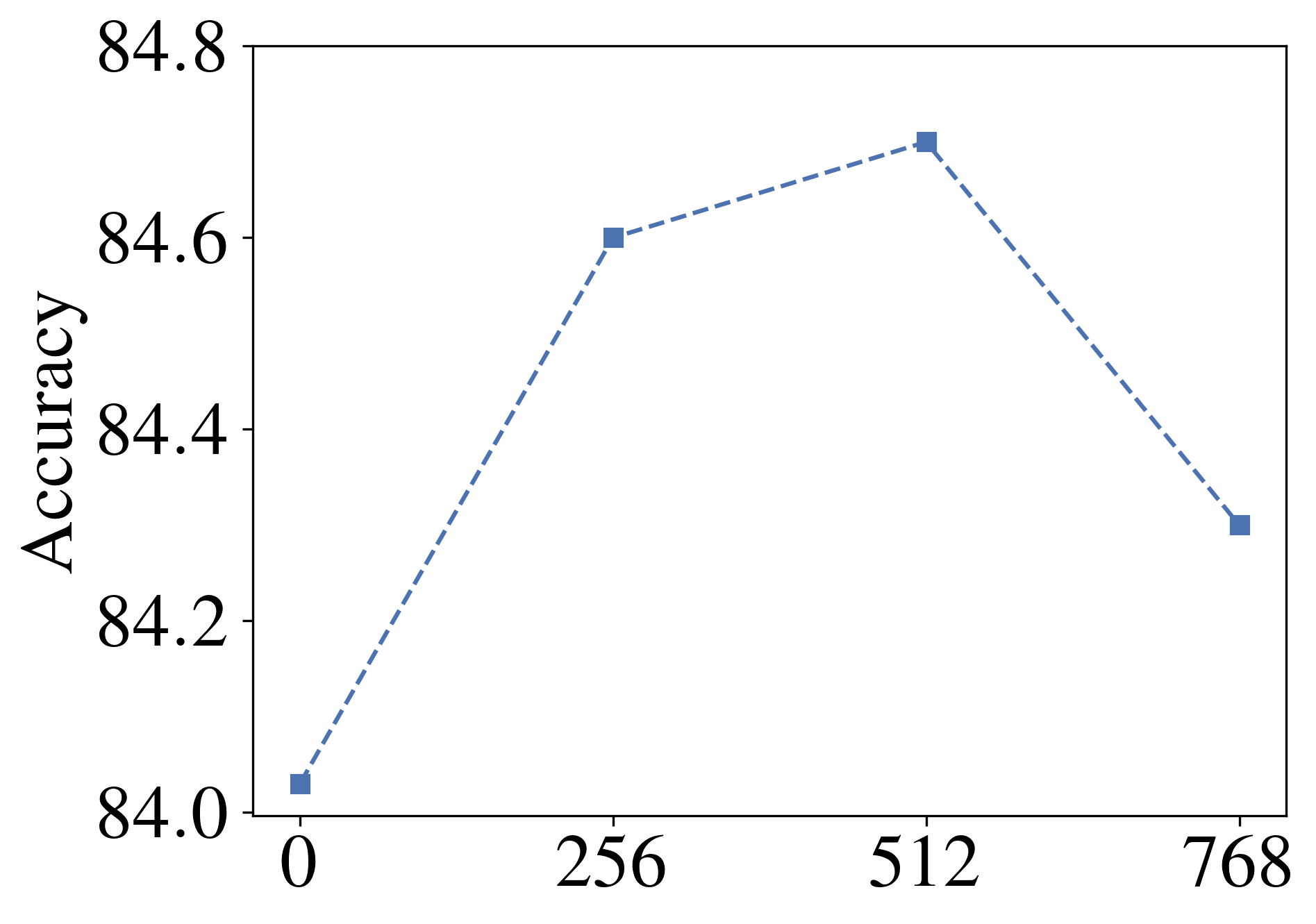}
    \caption{Shared dimension.}
    \label{fig:ablation:share-dim}
\end{subfigure}
\caption{Ablation study on MNLI. We report the average accuracy of MNLI-m and MNLI-mm. As default settings, we have expert dimension $768$, number of experts $4$, and shared dimension $512$.}
\end{figure*}

\subsection{Main Results}

Table~\ref{tab:glue-results} summarizes experimental results on the GLUE benchmark. Notice that our method outperforms all of the baseline methods in $6/7$ tasks.
In general task-agnostic distillation behaves better than task-specific algorithms because of the pre-training stage. For example, the best-performing task-specific method (BERT-of-Theseus) has a $68.2$ accuracy on the RTE dataset, whereas accuracy of MiniLMv2 and TinyBERT are greater than $72$. Using the proposed method, \ours obtains a $74.0$ accuracy on RTE without any pre-training, indicating the effectiveness of the MoE architecture.
We remark that \ours behaves on par or better than the vanilla BERT-base model in all of the tasks. This shows that there exists redundancy in pre-trained language models, which paves the way for model compression.

Table~\ref{tab:squad-results} summarizes experimental results on two question answering datasets: SQuAD v1.1 and SQuAD v2.0. Notice that \ours significantly outperforms all of the baseline methods in terms of both evaluation metrics: exact match (EM) and F1.
Similar to the findings in Table~\ref{tab:glue-results}, task-agnostic distillation methods generally behave better than task-specific ones. For example, PKD has a $69.8$ F1 score on SQuAD 2.0, while performance of MiniLMv1 and MiniLMv2 is over $76$.
Using the proposed MoE architecture, performance of our method exceeds both task-specific and task-agnostic distillation, e.g., the F1 score of \ours on SQuAD 2.0 is $76.8$, which is $7.0$ higher than PKD (task-specific) and $0.4$ higher than MiniLMv2 (task-agnostic).

\newcolumntype{C}{@{\hskip4pt}c@{\hskip4pt}}
\begin{table}[t!]
\centering
\begin{tabular}{l|CCCC}
\toprule
& \textbf{RTE} & \textbf{MNLI} & \textbf{SQuAD v2.0} \\ 
& Acc & m/mm & EM/F1 \\ \midrule
\ours & 74.0 & 84.5/84.8 & 73.6/76.8 \\ 
\quad~~ -distill & 73.3 & 83.2/84.0 & 72.5/76.0 \\
\bottomrule
\end{tabular}
\caption{Effectiveness of layer-wise distillation.}
\label{tab:ablation:distill}
\end{table}

\subsection{Ablation Study}

\vspace{0.1in} \noindent
\textbf{Expert dimension.}
We examine the affect of expert dimension, and experimental results are illustrated in Figure~\ref{fig:ablation:expert-dim}. As we increase the dimension of the experts, model performance improves. This is because of the increased model capacity due to a larger number of effective parameters.

\vspace{0.1in} \noindent
\textbf{Number of experts.}
Figure~\ref{fig:ablation:num-experts} summarizes experimental results when we modify the number of experts. As we increase the number of experts, model performance improves because we effectively enlarge model capacity. We remark that having only one expert is equivalent to compressing the model without incorporating MoE. In this case performance is unsatisfactory because of the limited representation power of the model.

\vspace{0.1in} \noindent
\textbf{Shared dimension.}
Recall that we share important neurons among the experts when adapting the FFNs. In Figure~\ref{fig:ablation:share-dim} we examine the effect of varying the number of shared neurons. Notice that sharing no neurons yields the worst performance, indicating the effectiveness of the sharing strategy. Also notice that performance of sharing all the neurons is also unsatisfactory. We attribute this to the lack of diversity among the experts.

\begin{figure*}
\centering
\begin{minipage}[b]{0.72\textwidth}
\begin{subfigure}{0.33\textwidth}
    \centering
    \includegraphics[width=1.0\textwidth]{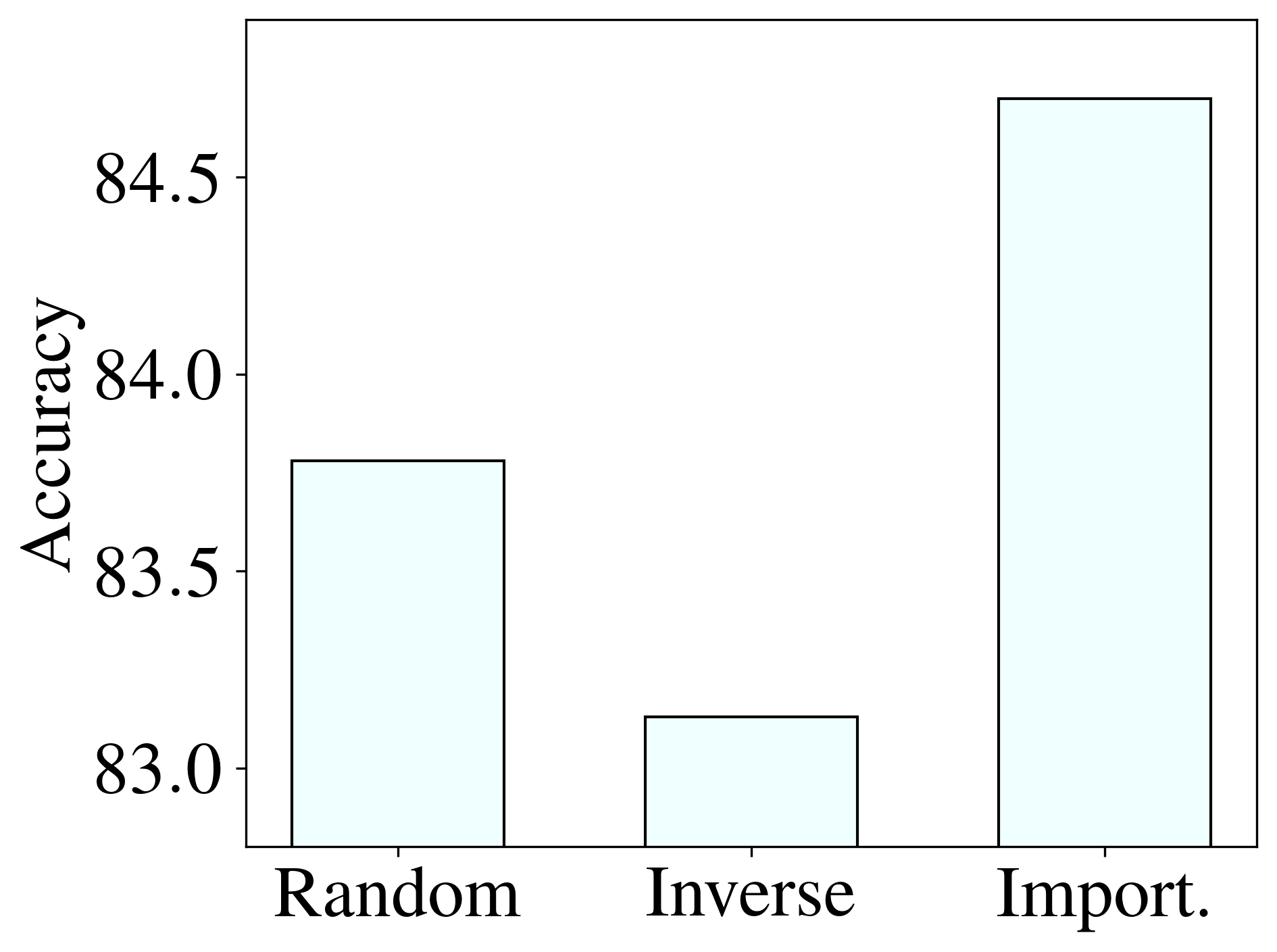}
    \caption{Adaptation methods.}
    \label{fig:ablation:split-method}
\end{subfigure}%
\begin{subfigure}{0.33\textwidth}
    \centering
    \includegraphics[width=1.0\textwidth]{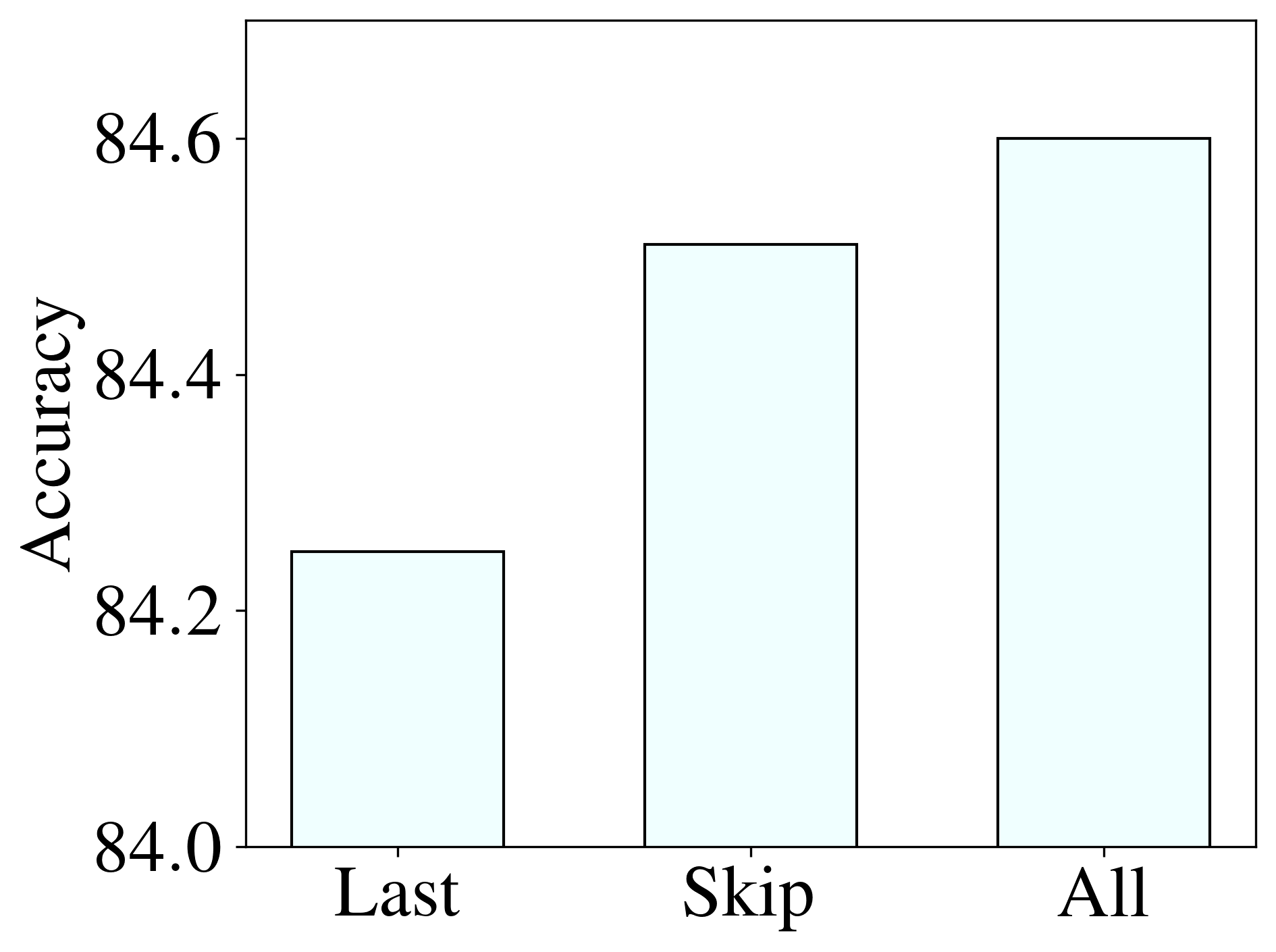}
    \caption{Distillation methods.}
    \label{fig:ablation:distill-method}
\end{subfigure}%
\begin{subfigure}{0.33\textwidth}
    \centering
    \includegraphics[width=1.0\textwidth]{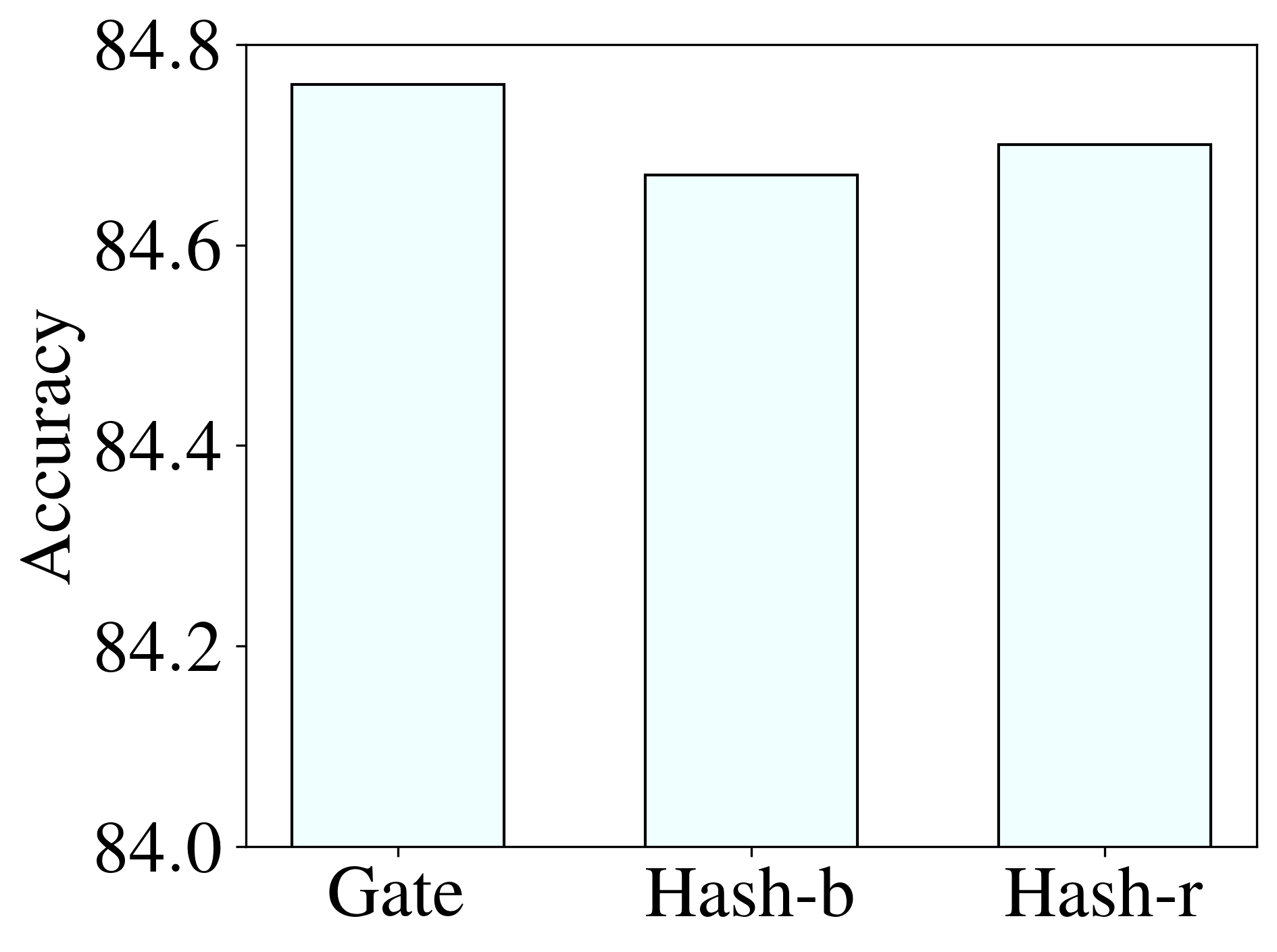}
    \caption{Routing methods in MoE.}
    \label{fig:ablation:moe-method}
\end{subfigure}
\caption{Experimental results of model variants on MNLI (average of m and mm). Our methods are denoted \textit{Import}, \textit{All} and \textit{Hash-r} in the subfigures, respectively.}
\end{minipage} \hfill
\begin{minipage}[b]{0.25\textwidth}
\centering
\includegraphics[width=1.0\linewidth]{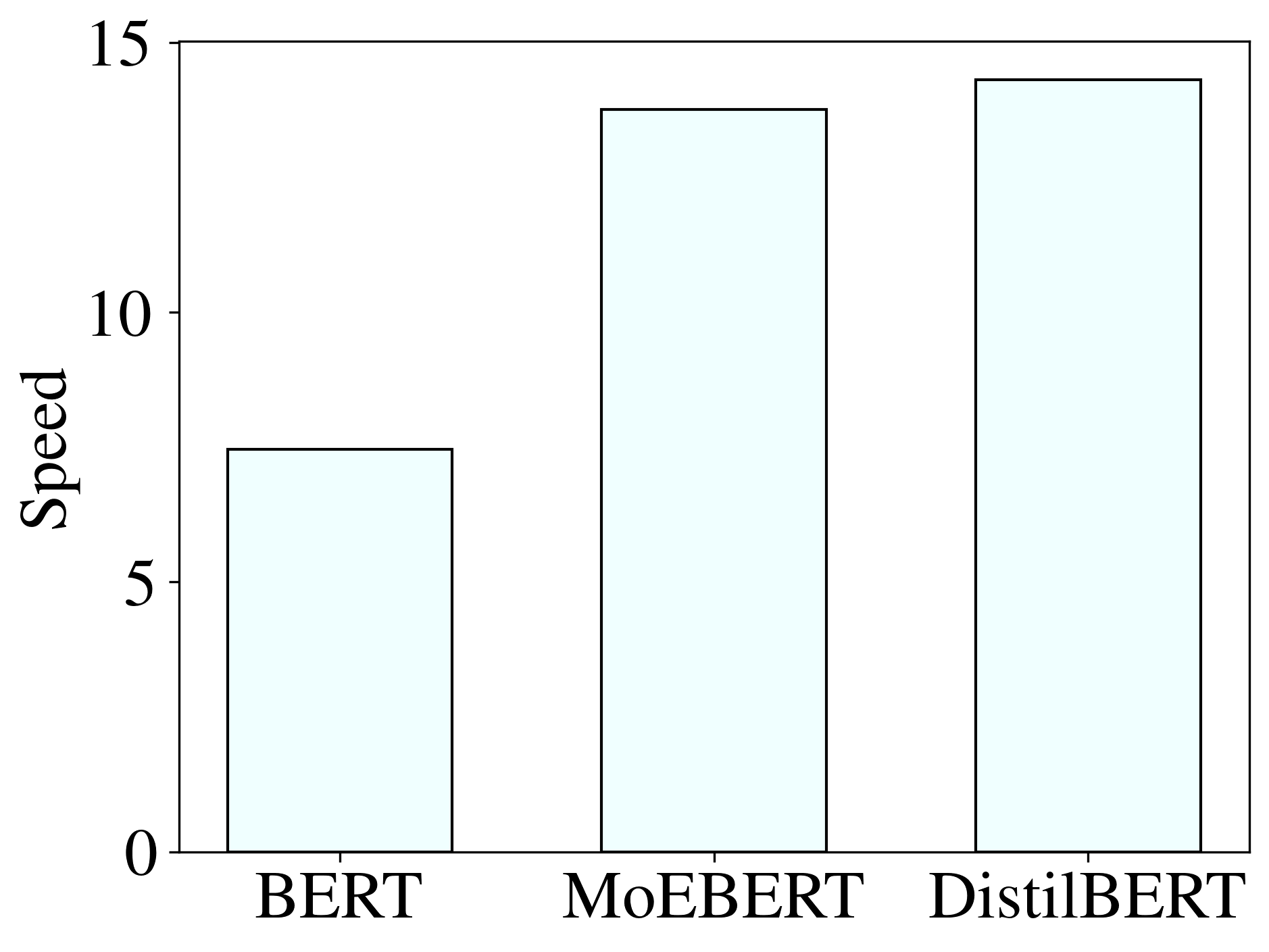}
\caption{Inference speed (examples/second, CPU) on the SST-2 dataset.}
\label{fig:ablation:inference}
\end{minipage}
\end{figure*}

\subsection{Analysis}

\textbf{Effectiveness of distillation.}
After adapting the FFNs in the pre-trained BERT-base model into experts, we train \ours using layer-wise knowledge distillation. In Table~\ref{tab:ablation:distill}, we examine the effectiveness of the proposed distillation method. We show experimental results on RTE, MNLI and SQuAD v2.0, where we remove the distillation and directly fine-tune the adapted model. Results show that by removing the distillation module, model performance significantly drops, e.g., accuracy decreases by $0.7$ on RTE and the exact match score decreases by $1.1$ on SQuAD v2.0.

\vspace{0.1in} \noindent
\textbf{Effectiveness of importance-based adaptation.}
Recall that we adapt the FFNs in BERT-base into experts according to the neurons' importance scores (Eq.~\ref{eq:importance}). We examine the method's effectiveness by experimenting on two different strategies: randomly split the FFNs into experts (denoted \textit{Random}), and adapt (and share) the FFNs according to the inverse importance, i.e., we share the neurons with the smallest scores (denoted \textit{Inverse}). Figure~\ref{fig:ablation:split-method} illustrated the results. Notice that performance significantly drops when we apply random splitting compared with \textit{Import} (the method we use). Moreover, performance of \textit{Inverse} is even worse than random splitting, which further demonstrates the effectiveness of the importance metric.

\vspace{0.1in} \noindent
\textbf{Different distillation methods.}
\ours is trained using a layer-wise distillation method (Eq.~\ref{eq:distil-loss}), where we add a distillation loss to every intermediate layer (denoted \textit{All}). We examine two variants: (1) we only distill the hidden states of the last layer (denoted \textit{Last}); (2) we distill the hidden states of every other layer (denoted \textit{Skip}). Figure~\ref{fig:ablation:distill-method} shows experimental results. We see that only distilling the last layer yields unsatisfactory performance; while the \textit{Skip} method obtains similar results compared with \textit{All} (the method we use).

\vspace{0.1in} \noindent
\textbf{Different routing methods.}
By default, we use a random hashing strategy (denoted \textit{Hash-r}) to route input tokens to experts \citep{roller2021hash}. That is, each token in the vocabulary is pre-assigned to a random expert, and this assignment remains the same during training and inference. We examine other routing strategies:
\begin{enumerate}
    \item We employ sentence-based routing with a trainable gate as in Eq.~\ref{eq:moe-gate} (denoted \textit{Gate}). Note that in this case, token representations in a sentence are averaged to compute the sentence representation, which is then fed to the gating mechanism for routing.
    Such a sentence-level routing strategy can significantly reduce communication overhead in MoE models. Therefore, it is advantageous for inference compared with other routing methods.
    \item We use a balanced hash list \citep{roller2021hash}, i.e., tokens are pre-assigned to experts according to frequency, such that each expert receives approximately the same amount of inputs (denoted \textit{Hash-b}).
\end{enumerate}
From Figure~\ref{fig:ablation:moe-method}, we see that all the methods yield similar performance. Therefore, \ours is robust to routing strategies.

\vspace{0.1in} \noindent
\textbf{Inference speed.}
We examine inference speed of BERT, DistilBERT and \ours on the SST-2 dataset, and Figure~\ref{fig:ablation:inference} illustrates the results. Note that for \ours, we use the sentence-based gating mechanism as in Figure~\ref{fig:ablation:moe-method}.
All the methods are evaluated on the same CPU, and we set the maximum sequence length to $128$ and the batch size to $1$.
We see that the speed of \ours is slightly slower than DistilBERT, but significantly faster than BERT.
Such a speed difference is because of two reasons.
First, the gating mechanism in \ours causes additional inference latency.
Second, DistilBERT develops a shallower model, i.e., it only has $6$ layers instead of $12$ layers; whereas \ours is a narrower model, i.e., the hidden dimension is $768$ instead of $3072$.

\begin{table}[t!]
\centering
\begin{tabular}{l|CCC}
\toprule
& \textbf{RTE} & \textbf{MNLI-m} & \textbf{MNLI-mm} \\ 
& Acc & Acc & Acc \\ \midrule
BERT\textsubscript{large} & 71.1 & \textbf{86.3} & 86.2 \\ \hline
MoEBERT\textsubscript{large} & \textbf{72.2} & \textbf{86.3} & \textbf{86.5} \\
\bottomrule
\end{tabular}
\caption{Distilling BERT-large on RTE and MNLI.}
\label{tab:ablation:bert-large}
\end{table}

\vspace{0.1in} \noindent
\textbf{Compressing larger models.}
Task-specific distillation methods do not require pre-training. Therefore, these methods can be easily applied to other model architectures and sizes beyond BERT-base. 
We compress the BERT-large model. Specifically, we adapt the FFNs in BERT-large (with hidden dimension $4096$) into four experts, such that each expert has hidden dimension $1024$. We share the top-$512$ neurons among experts according to the importance score. After compression, the number of effective parameters is reduces by half. Table~\ref{tab:ablation:bert-large} demonstrates experimental results on RTE and MNLI. We see that similar to the findings in Table~\ref{tab:glue-results}, \ours behaves on par or better than BERT-large in all of the experiments.

\section{Conclusion}
\label{sec:conclusion}

We present MoEBERT, which uses a Mixture-of-Experts structure to distill pre-trained language models. Our proposed method can speedup inference by adapting the feed-forward neural networks (FFNs) in a pre-trained language model into multiple experts. Moreover, the proposed method largely retains model capacity of the pre-trained model. This is in contrast to existing approaches, where the representation power of the compressed model is limited, resulting in unsatisfactory performance.
To adapt the FFNs into experts, we adopt an importance-based method, which identifies and shares the most important neurons in a FFN among the experts. We further propose a layer-wise task-specific distillation algorithm to train \ours. 
We conduct systematic experiments on natural language understanding and question answering tasks. Results show that the proposed method outperforms existing distillation approaches.

\section*{Ethical Statement}

This paper proposes MoEBERT, which uses a Mixture-of-Experts structure to increase model capacity and inference speed. We demonstrate that MoEBERT can be used for model compression. Experiments are conducted by fine-tuning pre-trained language models on natural language understanding and question answering tasks. In all the experiments, we use publicly available data and models, and we build our algorithms using public code bases.
We do not find any ethical concerns.

\bibliography{anthology,custom}
\bibliographystyle{acl_natbib}

\clearpage
\appendix
\section{Dataset details}
\label{app:dataset}

Statistics of the GLUE benchmark is summarized in Table~\ref{tab:glue}. 
Statistics of the question answering datasets (SQuAD v1.1 and SQuAD v2.0) are summarized in Table~\ref{tab:squad}.

\begin{table}[h!]
    \centering
    \begin{tabular}{l|cc}
        \toprule
        &  \#Train & \#Validation \\ \hline
        SQuAD v1.1 & 87,599 & 10,570 \\
        SQuAD v2.0 & 130,319 & 11,873 \\
        \bottomrule
    \end{tabular}
    \caption{Statistics of the SQuAD dataset.}
    \label{tab:squad}
\end{table}

\section{Training Details}
\label{app:training}

We use Adam \citep{kingma2014adam} as the optimizer with parameters $(\beta_1, \beta_2)=(0.9, 0.999)$. We employ gradient clipping with a maximum gradient norm $1.0$, and we choose weight decay from $\{0,0.01,0.1\}$.
The learning rate is chosen from $\{1\times 10^{-5}, 2\times 10^{-5}, 3\times 10^{-5}, 4\times 10^{-5}\}$, and we do not use learning rate warm-up.
We train the model for $\{3,4,5,10\}$ epochs with a batch size chosen from $\{8,16,32,64\}$.
The weight of the distillation loss $\lambda\textsubscript{distil}$ is chosen from $\{1,2,3,4,5\}$.

Hyper-parameters for distilling BERT-base is summarized in Table~\ref{tab:hyper-parameters}.
We use Adam \citep{kingma2014adam} as the optimizer with parameters $(\beta_1, \beta_2)=(0.9, 0.999)$. We employ gradient clipping with a maximum gradient norm $1.0$. We do not use learning rate warm-up.
For the GLUE benchmark, we use a maximum sequence length of $512$ except MNLI and QQP, where we set the maximum sequence length to $128$. 
For the SQuAD datasets, the maximum sequence length is set to $384$.

\begin{table*}[b!]
	\begin{center}
		\begin{tabular}{l|l|c|c|c|c|c}
			\toprule 
			\bf Corpus &Task& \#Train & \#Dev & \#Test   & \#Label &Metrics\\ \midrule
			\multicolumn{6}{@{\hskip1pt}r@{\hskip1pt}}{Single-Sentence Classification (GLUE)} \\ \hline
			CoLA & Acceptability&8.5k & 1k & 1k & 2 & Matthews corr\\ \hline
			SST & Sentiment&67k & 872 & 1.8k & 2 & Accuracy\\ \midrule
			\multicolumn{6}{@{\hskip1pt}r@{\hskip1pt}}{Pairwise Text Classification (GLUE)} \\ \hline
			MNLI & NLI& 393k& 20k & 20k& 3 & Accuracy\\ \hline
            RTE & NLI &2.5k & 276 & 3k & 2 & Accuracy \\ \hline
			QQP & Paraphrase&364k & 40k & 391k& 2 & Accuracy/F1\\ \hline
            MRPC & Paraphrase &3.7k & 408 & 1.7k& 2&Accuracy/F1\\ \hline
			QNLI & QA/NLI& 108k &5.7k&5.7k&2& Accuracy\\ \midrule
			\multicolumn{5}{@{\hskip1pt}r@{\hskip1pt}}{Text Similarity (GLUE)} \\ \hline
			STS-B & Similarity &7k &1.5k& 1.4k &1 & Pearson/Spearman corr\\ \bottomrule
		\end{tabular}
	\end{center}
	\vskip -0.05in
	\caption{Summary of the GLUE benchmark.}
	\label{tab:glue}
\end{table*}

\begin{table*}[b!]
\centering
\begin{tabular}{l|ccccc}
\toprule
& lr & batch & epoch & decay & $\lambda\textsubscript{distill}$ \\ \midrule
RTE & $1\times 10^{-5}$ & $1 \times 8$ & $10$ & $0.01$ & $1.0$ \\
CoLA & $2\times 10^{-5}$ & $1 \times 8$ & $10$ & $0.0$ & $3.0$ \\
MRPC & $3\times 10^{-5}$ & $1 \times 8$ & $5$ & $0.0$ & $2.0$ \\
SST-2 & $2\times 10^{-5}$ & $2 \times 8$ & $5$ & $0.0$ & $1.0$ \\
QNLI & $2\times 10^{-5}$ & $4 \times 8$ & $5$ & $0.0$ & $2.0$ \\
QQP & $3\times 10^{-5}$ & $8 \times 8$ & $5$ & $0.0$ & $1.0$ \\
MNLI & $5\times 10^{-5}$ & $8 \times 8$ & $5$ & $0.0$ & $5.0$ \\
SQuAD v1.1 & $3\times 10^{-5}$ & $4 \times 8$ & $5$ & $0.01$ & $2.0$ \\
SQuAD v2.0 & $3\times 10^{-5}$ & $2 \times 8$ & $4$ & $0.1$ & $1.0$ \\
\bottomrule
\end{tabular}
\caption{Hyper-parameters for distilling BERT-base. From left to right: learning rate; batch size ($2\times 8$ means we use a batch size of $2$ and $8$ GPUs); number of training epochs; weight decay; and weight of the distillation loss.}
\label{tab:hyper-parameters}
\end{table*}

\end{document}